\documentclass{article}

\usepackage{algorithm}
\usepackage{algorithmic}
\usepackage{graphicx}
\usepackage{multirow}
\usepackage{float}
\usepackage{adjustbox}
\usepackage{subcaption}
\usepackage{csquotes}
\usepackage{amsmath}
\usepackage{amssymb}
\usepackage{PRIMEarxiv}

\usepackage[utf8]{inputenc} 
\usepackage[T1]{fontenc}    
\usepackage{hyperref}       
\usepackage{url}            
\usepackage{booktabs}       
\usepackage{amsfonts}       
\usepackage{nicefrac}       
\usepackage{microtype}      
\usepackage{lipsum}
\usepackage{fancyhdr}       
\usepackage{graphicx}       
\graphicspath{{media/}}     

\title{Causal Discovery and Classification Using Lempel-Ziv Complexity
}

\author{
 Dhruthi\\
  Department of Computer Science and Information Systems \\
  BITS Pilani K K Birla Goa Campus\\
  403726, Goa, India\\
  \texttt{f20231019@goa.bits-pilani.ac.in}  \\
   \And
  Nithin Nagaraj \\
  Complex Systems Programme \\
  National Institute of Advanced Studies, Indian Institute of Science Campus\\
  Bengaluru, 560012, Karnataka, India\\
  \texttt{nithin@nias.res.in} \\
  \And
   Harikrishnan N B\\
  Department of Computer Science and Information Systems \\
  BITS Pilani K K Birla Goa Campus\\
  403726, Goa, India\\
  \texttt{harikrishnannb@goa.bits-pilani.ac.in}  \\
}

\begin{document}
\maketitle

\begin{abstract}
Inferring causal relationships in the decision-making processes of machine learning algorithms is a crucial step toward achieving explainable Artificial Intelligence (AI). In this research, we introduce a novel causality measure and a distance metric derived from Lempel-Ziv (LZ) complexity. We explore how the proposed causality measure can be used in decision trees by enabling splits based on features that most strongly \textit{cause} the outcome. We further evaluate the effectiveness of the causality-based decision tree and the distance-based decision tree in comparison to a traditional decision tree using Gini impurity. While the proposed methods demonstrate comparable classification performance overall, the causality-based decision tree significantly outperforms both the distance-based decision tree and the Gini-based decision tree on datasets generated from causal models. This result indicates that the proposed approach can capture insights beyond those of classical decision trees, especially in causally structured data. Based on the features used in the LZ causal measure based decision tree, we introduce a causal strength for each features in the dataset so as to infer the predominant causal variables for the occurrence of the outcome. 
\end{abstract}


Incorporating causal based explanation to Machine Learning (ML) decision making process is an integral part of trustworthiness and model explanbaility~\cite{longo2024explainable}. In many scenario's a dataset dealt by an ML scientist would be observational data. Observational data are historical records of a phenomenon observed passively. An example of observational data would be  studying the link between smoking and lung cancer. Such a study includes the documentation of the historical recording of the smoking/non-smoking habits and health condition over time. It is essential to note that such studies do not ask people to start and quit smoking.   On the other hand, interventional data includes design experiments to study the effect of a specific variable through randomized control trials (RCT's)~\cite{stolberg2004randomized}.  For eg., in the COVID -19 vaccine trials, researchers do interventional study to establish the causal influence of a proposed treatment for a desired outcome through RCT's. In several cases, conducting RCT's is not feasible due to ethical reasons. So, it is essential to have robust methods that can give causal explanation from observational data. 
There are several methods used to detect causal directions from observational data. The methods like Granger Causality,~\cite{granger1969investigating} (GC), Transfer Entropy~\cite{schreiber2000measuring} (TE) make use of Norbert Wiener's notion of causality~\cite{wiener1956theory}. Wiener's definition of Causality states -- \emph{if a time series X causes a time series Y, then past values of X should contain information that help predict Y above and beyond the information contained in past values of Y alone}~\cite{kathpalia2021measuring}. The above definition of predictability as an indicative measure for causality is expanded to incorporate compressibility as they are related. The following methods uses compressibility as an causality indicative measure:  Compression Complexity Causality~\cite{kathpalia2019data}, ORIGO~\cite{budhathoki2018origo}, ERGO~\cite{vreeken2015causal}. ORIGO is a causal discovery method based on minimum description length principle (MDL). ORIGO infers that $X$ is likely a cause of $Y$ if compressing $X$ first, followed by compressing $Y$ given $X$, results in better compression than doing it in the reverse order~\cite{budhathoki2018origo}. ERGO relies on complexity estimates as proxies for Kolmogrov complexity instead lossless compression~\cite{vreeken2015causal}. In~\cite{pranay2021causal}, the authors introduced a penalty and efficacy formulation utilizing Lempel-Ziv ($LZ$) complexity and Effort to Compress (ETC) to develop a new causality measure. This method was evaluated on various datasets, including an autoregressive (AR) process, the Tuebingen datasets~\cite{mooij2016distinguishing}, and SARS-CoV-2 viral genome datasets. The performance of their approach was compared with that of state-of-the-art methods, specifically ORIGO and ERGO. The results from all experiments conducted in~\cite{pranay2021causal} demonstrated that the proposed $LZ$ and ETC-based causality measures perform competitively against these established techniques on synthetic benchmarks. 

The aim of this research is to develop novel causal direction identifying measure from univariate datasets, and incorporate in decision tree algorithm for developing a causally informed model. The steps undertaken to achieve the goals are as follows:
\begin{enumerate}
    \item Novel causal measure based on Lempel-Ziv ($LZ$) Complexity - Inspired by the efficacy of the compression complexity based causality measures, we propose a novel causality testing measure based on Lempel-Ziv complexity to check the direction of causation from univariate temporal/non-temporal data.  The proposed method differs significantly from the Lempel-Ziv and ETC-based causality measures introduced in~\cite{pranay2021causal}. We demonstrate a use case where our approach outperforms the causality measure suggested in the above paper. 
    \item Testing of the causality measure - Test the efficacy of the causality measure in real world and synthetic datasets.
    \item Integration of the proposed causal measure in ML decision tree model - we integrate the proposed measure into a decision tree as a splitting criterion, creating a tree whose decisions at each node are guided by our proposed causality measure.
    \item Novel distance metric derived from Lempel-Ziv complexity - introduced a novel distance metric derived from the Lempel-Ziv complexity measure, which is incorporated into the decision tree.
    \item Performance comparison - compare the two proposed methods with a decision tree that uses the classical {\it Gini impurity} on several datasets, including \emph{AR dataset, Iris, Breast Cancer, Voting, Car Evaluation, KRKPA7, Mushroom, Thyroid, Heart Disease.}
    \item Interpretabilty of our proposed model: We propose a {\it feature importance score} based on $LZ$ causal measure based decision tree. This score will rank the features based on its causal influence on the outcome variable.
    
\end{enumerate}

The sections in the paper are arranged as follows: Section~\ref{sec:proposed_method} describes the proposed method, the experiments on validating the $LZ$ based causal measure using synthetic dataset and incorporating $LZ$ based causal and distance measure with decision trees is highlighted in Section~\ref{sec:experiments}. In Section~\ref{sec:conclusion}, we provide the Conclusion. The appendix material can be accessed in Section~\ref{sec:appendix}.
\section{Proposed Method\label{sec:proposed_method}}
In this section, we describe the proposed causality and distance based measure derived from Lempel-Ziv complexity. The proposed causality measure has the following assumptions:

\begin{enumerate}
    \item Assumption 1: Cause happens either before the effect or they occur together. We don't consider retrocausal scenarios in this study.
    \item Assumption 2: There is no confounding variable that causes $X$ and $Y$, where $X$ and $Y$ are temporal/ non-temporal data. 
\end{enumerate} 
Consider two univariate dataset $X =[a,b,c,a,c,b,a,a \ldots]$ and $Y = [b,a,a,a,b,c,\ldots]$ which may or may not have temporal structure. We want to develop a measure using which we want to comment the direction causality. i.e., $X \rightarrow Y$ or $Y\rightarrow X$. We want the measure to have the following properties:
\begin{itemize}
    \item If $X \rightarrow Y$, then the grammar of $X$ constructed over real time has patterns that better explain $Y$. The extra penalty incurrred by explaining $Y$  using the generated grammar of $X$  denoted as $LZ-P_{X\rightarrow Y}$ is less compared to the penalty incurred by explaining $X$ using the generated grammar of $Y$ denoted as $LZ-P_{Y\rightarrow X}$. For this case, $LZ-P_{X\rightarrow Y} < LZ-P_{Y\rightarrow X}$. The inequality will be reversed if $Y \rightarrow X$.
    \item Explaining $X$ using the real time generated grammar of $X$ denoted as $LZ-P_{X\rightarrow X}$ should give zero penalty. This follows from the assumption 1.  
\end{itemize}

We now describe the construction of the real time grammar of a temporal/ non-temporal data for the proposed causality measure.
\subsection{Causality Measure based on Lempel-Ziv Complexity}\label{sec:defnlzp}
\subsubsection{Definition}
Given two symbolic sequences, $x$ = $x_0x_1x_2....x_m$ and $y$ = $y_0y_1y_2....y_n$, such that $0 \leq x_k \leq m-1$ and $0 \leq y_k \leq n-1$, we define $LZ-P_{x\rightarrow y}$ as the penalty incurred by explaining $y$ using the real time grammar of $x$. The algorithm for calculating the penalty is described in Algorithm~\ref{alg:lz-p}.
For example, consider the two strings $x$ = ``$ 101110$" and $y$ =``$ 110111$". Then $LZ-P_{x\rightarrow y}$ is calculated according to the following steps:{
\begin{itemize}
 \item \textbf{Step 0}: We consider the two strings $x$ and $y$, and construct their respective grammar sets $G_x$ and $G_y$ respectively using the Lempel-Ziv algorithm. We monitor the overlap at each stage, which can be thought to represent the extent to which the current and previous values of $x$ influence $y$.
    \[
    x ~=~ \text{101110}, \quad y~=~\text{110111},\quad \text{overlap} = 0,  \quad G_x = \phi, \quad G_{y}=\phi.
    \]
    \item \textbf{Step 1}: Since we are inferring the strength of causation from string $x$ to string $y$, we start with the selection of the smallest substring of $x$, starting from the first index, that is not already present in its grammar set. Since the grammar set is empty, `1' is selected and added to $G_x$.
    \[
    x ~=~ \underline{\textbf{1}}\text{01110}, \quad \text{overlap} = 0,  \quad G_x = \{`1\text{'}\}, \quad G_{y}=\phi.
    \]
        \item \textbf{Step 2}: Similarly, for string $y$, the substring `1' is selected. Since this substring is already present in $G_x$, overlap is incremented. We can interpret this as the `1' present in $x$ inducing its occurrence in $y$.
    \[   y~=~\underline{\textbf{1}}\text{10111}, \quad  \text{overlap} = 1, \quad G_x = \{ `1\text{'}\}, \quad G_{y}=\{ `1\text{'} \}.
    \]
    \item \textbf{Step 3}: The process is now repeated. The starting index of the subsequent substring must immediately follow the terminal index of the preceding substring, so that no information is lost. Hence `0' is selected and added to the set $G_x$.
       \begin{center} $ x ~=~ \text{1}\underline{\textbf{0}}\text{1110}, \quad \text{overlap} = 1,  \quad G_x = \{`1\text{'},`0\text{'}\}, \quad G_{y}=\{`1\text{'}\}.$ \end{center}  
        \item \textbf{Step 4}: As in Step 3, we consider the smallest  subsequently occuring substring, `1'. Since it is already present in $G_y$, we consider the substring `10' instead. `10' is not present in $G_y$ and is  hence added to it, and since it is not present in $G_x$, overlap is not incremented. 
    \[
   y~=~\text{1}\underline{\textbf{10}}\text{111}, \quad  \text{overlap} = 1, \quad G_x = \{`1\text{'},`0\text{'}\}, \quad G_{y}=\{ `1\text{'},`10\text{'} \}.
    \]
    \item \textbf{Step 5}:The substring `1' is already present in $G_x$. The substring `11' is selected and added to $G_x$, as in Step 3.
     \[
    x ~=~ \text{10}\underline{\textbf{11}}\text{10}, \quad \text{overlap} = 1,  \quad G_x = \{`1\text{'},`0\text{'},`11\text{'}\}, \quad G_{y}=\{`1\text{'},`10\text{'}\}.
    \]
        \item \textbf{Step 6}: The substring `1' is already present in $G_y$. The substring `11' is selected and added to $G_y$. Since it is already present in $G_x$, overlap is incremented.
    \[
   y~=~\text{110}\underline{\textbf{11}}\text{1}, \quad  \text{overlap} = 2, \quad G_x = \{`1\text{'},`0\text{'},`11\text{'}\}, \quad G_{y}=\{ `1\text{'},`10\text{'}, `11\text{'}\}.
    \]
    \item \textbf{Step 7}: The substring `10' is not present in $G_x$, and is hence added to $G_x$.
    \[
   x~=~\text{1011}\underline{\textbf{10}}, \quad  \text{overlap} = 2, \quad G_x = \{`1\text{'},`0\text{'},`11\text{'}\,`10\text{'}\}, \quad G_{y}=\{ `1\text{'},`10\text{'}, `11\text{'}\}.
    \]
    
    Since there are no more substrings left in either string that can be added to the respective grammar sets, the process stops here, and $LZ-P_{x\rightarrow y}$ is found to be ($|G_y|$ - overlap), which is 1. In the same manner, $LZ-P_{y\rightarrow x}$ is also found to be 1. \\
    If $LZ-P_{x\rightarrow y}$ < $LZ-P_{y\rightarrow x}$, we can say that $x$ causes $y$. Similarly, if $LZ-P_{y\rightarrow x}$ < $LZ-P_{x\rightarrow y}$, we conclude that $y$ causes $x$. In this case, since the penalties are equal, the flow of information between $y$ and $x$ is bidirectional and we decide the direction of causation by a coin flip.
    
\end{itemize}

\begin{algorithm}[!ht]
\caption{Calculation of Penalty of a string $y$ of length $n$ given another string x of length $m$} 
\label{alg:lz-p}
\begin{algorithmic}[1]
    \STATE \textbf{Input:} String $y$ of length $n$, String $x$ of length $m$.
    \STATE \textbf{Output:} Penalty of using $x$ to compress $y$
    \STATE Initialize set $G_{x} \gets \emptyset$  \COMMENT{Empty grammar set}
    \STATE Initialize set $G_{y} \gets \emptyset$  \COMMENT{Empty grammar set}
    \STATE Initialize $i_x \gets 1$
    \STATE Initialize $i_y \gets 1$
    \STATE Initialize $overlap \gets 0$
    \WHILE{$i_y \leq n$}
    \IF{$i_x \leq m$}
        \STATE Set $j_x \gets i_x$
        \WHILE{$x[i_x:j_x] \in G_x$  \AND $j_x < m$ }
            \STATE Increase $j_x$ by 1
        \ENDWHILE
        \STATE Add substring $x[i:j]$ to $G_{x}$
    \ENDIF
        \STATE Set $j_y \gets i_y$
        \WHILE{$y[i_y:j_y] \in G_y$ \AND $j < n$ }
            \STATE Increase $j_y$ by 1
        \ENDWHILE
        \IF {$y[i_y:j_y] \in G_x$}
            \STATE Increase $overlap$ by 1
        \ENDIF
        \STATE Add $y[i_y:j_y]$ to $G_y$
        \STATE Set $i_x \gets j_x+1$
        \STATE Set $i_y \gets j_y+1$
        
    \ENDWHILE
    \STATE \textbf{return} $|G_y|-overlap$
\end{algorithmic}
\end{algorithm}
In the upcoming section~\ref{sec:experiments}, we test the correctness of the proposed causality measure for determining the direction of causation from univariate datasets.
\subsection{A Distance metric based on Lempel-Ziv Complexity\label{sec:lz-distance-measure}}
In this section, we define the a distance metric derived from the grammar of two  symbolic sequences. The grammar construction is based on Lempel-Ziv algorithm~\cite{sayood2017introduction}.
Given two symbolic sequences, $x$ = $x_0x_1x_2....x_m$ and $y$ = $y_0y_1y_2....y_n$, such that $0 \leq x_k \leq m-1$ and $0 \leq y_k \leq n-1$, their grammars $G_x$ and $G_y$ respectively can be encoded using the Lempel-Ziv Algorithm, as shown in Algorithm \ref{alg:lz}. The Lempel-Ziv distance between two sets $G_x$ and $G_y$ is
\begin{equation}
d_{LZ}(G_x,G_y) = |G_x/G_y| + |G_y/G_x|,
\end{equation} where $|G|$ represents the cardinality of a set $G$.
The proof of the given measure being a distance metric on the metric space of the power set of set of all symbolic sequences of all finite lengths is given in Appendix~\ref{sec:appendix_proof}. Decision trees utilizing this metric select thresholds depending on the symbolic sequence structure, and are thus sensitive to the permutation of training data.
\begin{algorithm}

\caption{Calculation of Grammar of a String Using Lempel-Ziv Complexity}
\label{alg:lz}
\begin{algorithmic}[1]
    \STATE \textbf{Input:} String $x$ of length $n$ 
    \STATE \textbf{Output:} Grammar of the string, $G_x$
    \STATE Initialize set $G_x \gets \emptyset$  \COMMENT{Empty grammar set}
    \STATE Initialize $i \gets 1$
    \WHILE{$i \leq n$}
        \STATE Set $j \gets i$
        \WHILE{$x[i:j] \in G_x$ \AND $j < n$}
            \STATE Increase $j$ by 1
        \ENDWHILE
        \STATE Add substring $x[i:j]$ to $G_x$
        \STATE Set $i \gets j+1$
    \ENDWHILE
    \STATE \textbf{return} $G_x$
\end{algorithmic}
\end{algorithm}
\subsection{Applicability in Decision Trees}
In this section, we highlight on the utilization of the proposed causality and distance measure as a splitting criteria in decision tree thereby yielding (a) Lempel-Ziv Causal Measure based decision tree, (b) Lempel-Ziv distance metric based decision tree.

\subsubsection{Utilization as a splitting criteria}
To split a binary tree at a given node, a feature and threshold are selected based on minimizing $LZ$-based causality measure between a feature string (for a particular chosen threshold) and target string. Each feature and target value is transformed into symbolic sequences: elements in the feature column are represented as 0 if below the threshold, and 1 if above. In the target symbolic sequence $y$, a given label $l$ is represented as 1 and all others as 0. The feature and threshold selection is done as per the following equation: 

\begin{equation}
    (best~feature,best~threshold) = \arg\min_{feature,target,label}(LZ-P_{feature\rightarrow target}).
\end{equation}
A similar approach is utilized to implement a $LZ$-distance based decision tree, now replacing the $LZ$-P causality meaure with the $LZ$-based distance measure as defined in Section~\ref{sec:lz-distance-measure}. Once the symbolic sequences are found, algorithm~\ref{alg:lz} is utilized to find the grammar of the symbolic sequences, represented by $G_{feature}$ and $G_{target}$ respectively. The feature, threshold pair with minimum distance with respect to the target label is selected, based on the following equation.
\begin{equation}
(best~feature,best~threshold) = \arg\min_{feature,target,label}(d_{LZ}(G_{feature},G_{target})).
\end{equation}

\subsubsection{Causal Strength of a feature on a target}
We propose a method of ranking of the strength of causation of different attributes on a particular label based on the causal decision tree. Consider a decision tree with $m$ nodes, and $d_i$ denotes the depth of the $i$th node. The causal strength $s_j$ for a feature $j$ to the target is given by the following formulation:
\begin{equation}
    s_j = \sum_{i=0}^{m} a_{ij} \cdot 2^{-d_i}, \quad \text{where} \quad a_{ij} = 
\begin{cases} 
      1 & \text{if } j \text{ is the splitting feature at the } i\text{th node}, \\
      0 & \text{otherwise}
   \end{cases}
\end{equation}

These scores can then be normalized. This provides a relative ranking of causal influence of the features on the target attribute. 



\section{Experiments\label{sec:experiments}}
In this section, we first demonstrate the efficacy of the proposed causal measure in successfully determining the direction of causation for coupled Auto Regressive (AR) processes, coupled logistic map as described in section~\ref{sec:coupled_AR_process} and section~\ref{sec:coupled_logistic_map} respectively. We also compare the performance of proposed method with penalty measure defined in~\cite{pranay2021causal} for real world dataset obtained from Tuebingen cause-effect pairs repository\footnote{https://webdav.tuebingen.mpg.de/cause-effect/} as mentioned in Section~\ref{sec:tubingen_cause-effect_pairs}. The integration of the proposed casual measure and distance based measure with decision trees and its performance comparison with decision tree based on gini impurity for the following datasets \emph{AR dataset, Iris, Breast Cancer, Voting, Car Evaluation, KRKPA7, Mushroom, Thyroid, Heart Disease}.
\subsection{Simple causation in AR processes of various orders\label{sec:coupled_AR_process}}
Data was generated from a coupled auto regressive process of order $p = 1, 5, 20, 100$. The governing equation for $X(t)$ and $Y(t)$, where both are timeseries data are as follows:
\begin{eqnarray}
    X(t) = aX(t-1) + \eta Y(t-p) + \epsilon_{X,t}, \\
    Y(t) = bY(t-1) + \epsilon_{Y,t}. 
\end{eqnarray}
The parameters were set to be $a = 0.9$, $b = 0.9$ and coupling coefficient ($\eta$) was varied from $\eta=0$  to $\eta = 1$  with step size = $0.1$. Noise terms $\epsilon_{X,t} = \nu~ e_1$ and $\epsilon_{Y,t} = \nu ~e_2$ where $e_1$ and $e_2$ are sampled from the standard normal distribution and noise intensity $\nu = 0.03$. The process was simulated for $t = 1000$ time steps, with the first 500 transients discarded, resulting in a time series of length 2000 for both $X(t)$ and $Y(t)$. This procedure was repeated for $n = 1000$ trials for each value of the coupling coefficient. Figure~\ref{fig_ar_1_5_20_100_var_LZ_penalty_coupling_coefficient} a, b, c and d represents the average (across 1000 trials for each coupling coefficient) $LZ$ penalty from $X(t)$ to $Y(t)$ and $Y(t)$ to $X(t)$ averaged across $1000$ independent random trials for each coupling coefficient corresponding to AR-1, AR-5, AR-20 and AR-100 respectively. The error bars in the plot represents the standard deviation. The proposed measure captures the correction direction of causality from $Y(t)$ to $X(t)$. 

\begin{figure}[!ht]

    \centering
  
    \begin{subfigure}[]{0.45\textwidth}
        \centering
        \includegraphics[width=\linewidth]{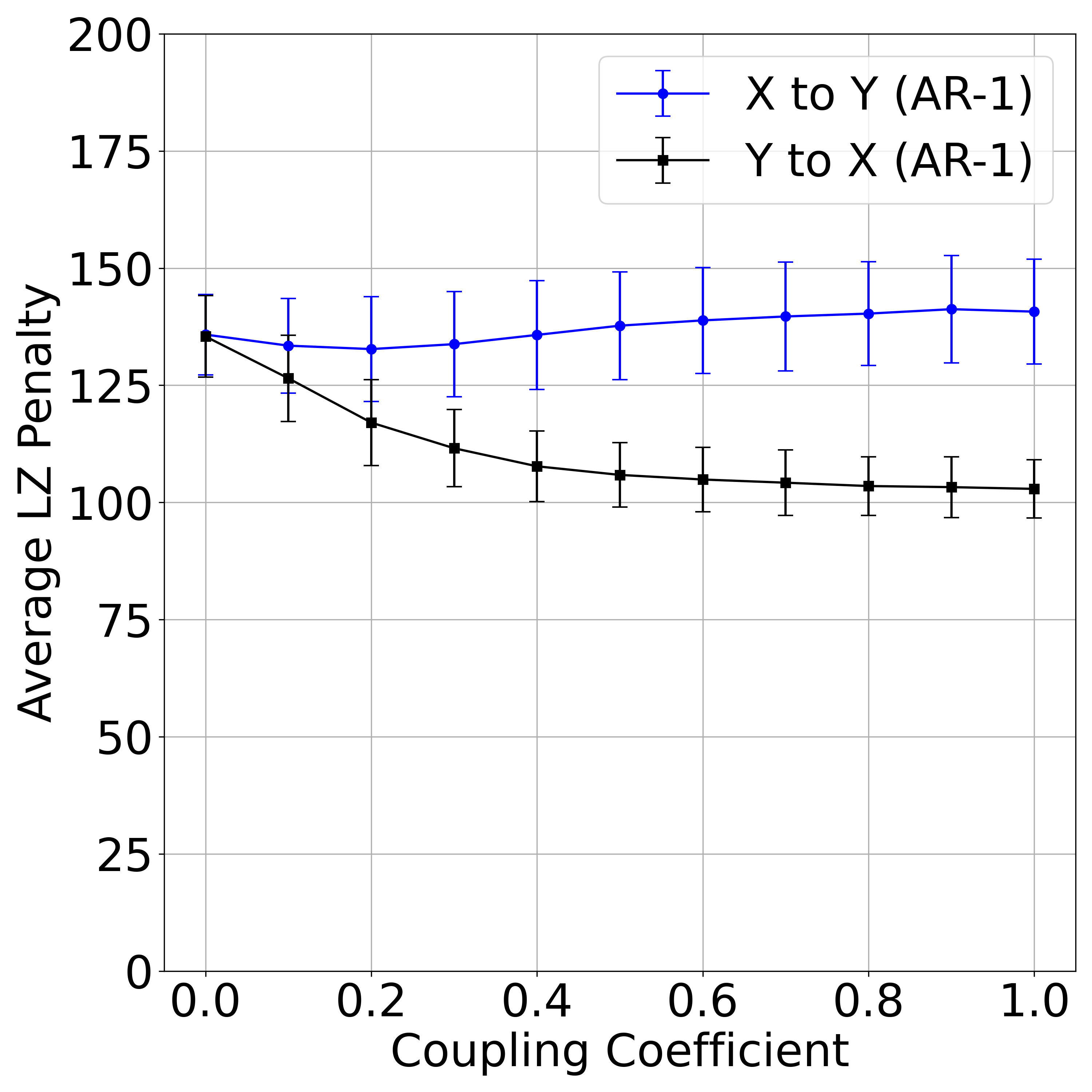}
        ~~(a)\label{fig_ar_1_LZ_penalty_coupling_coefficient}
    \end{subfigure}
    \hfill
    \begin{subfigure}[]{0.45\textwidth}
        \centering
        \includegraphics[width=\linewidth]{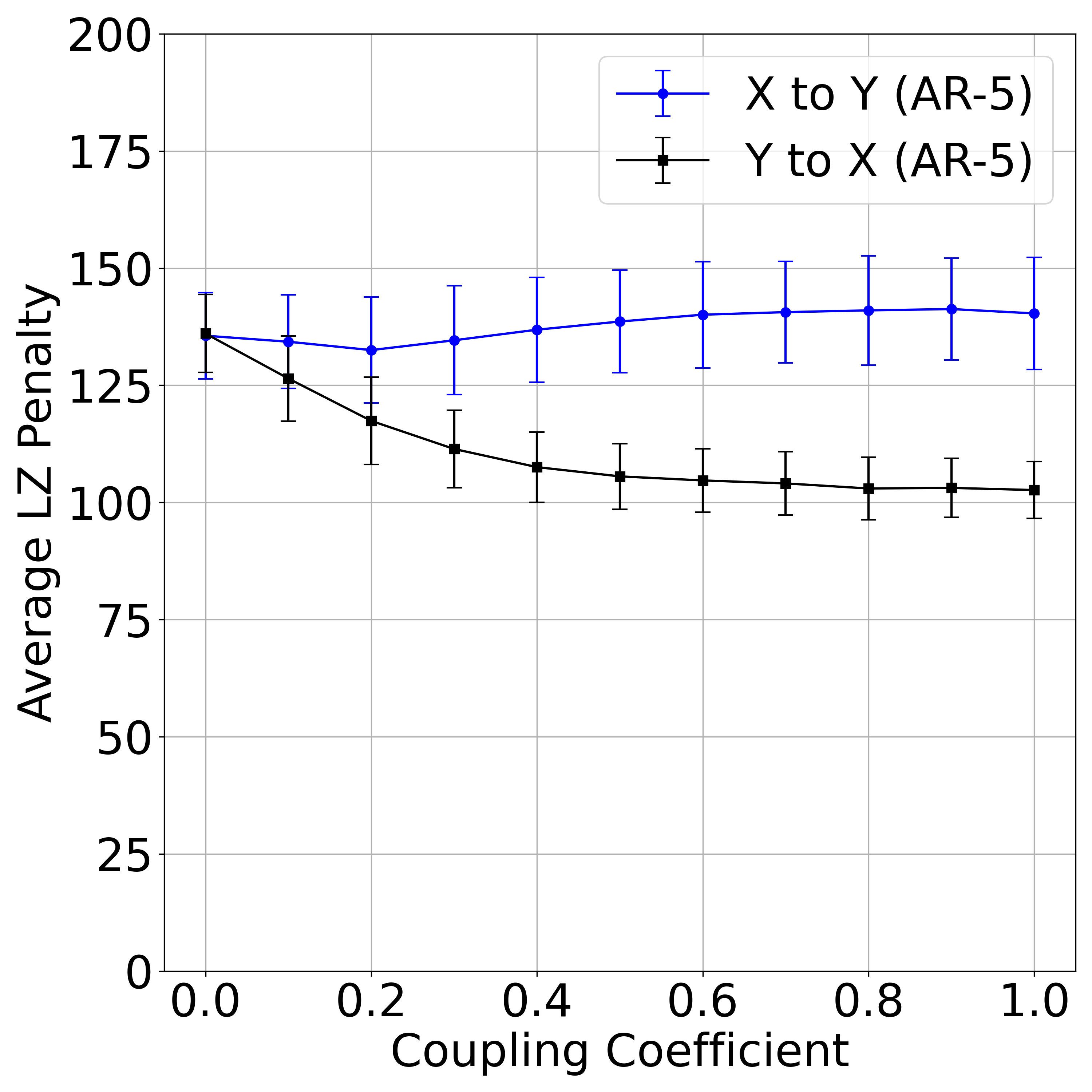}
        ~~(b)\label{fig_ar_5_var_LZ_penalty_coupling_coefficient}
    \end{subfigure}
    
    \begin{subfigure}[]{0.45\textwidth}
        \centering
        \includegraphics[width=\linewidth]{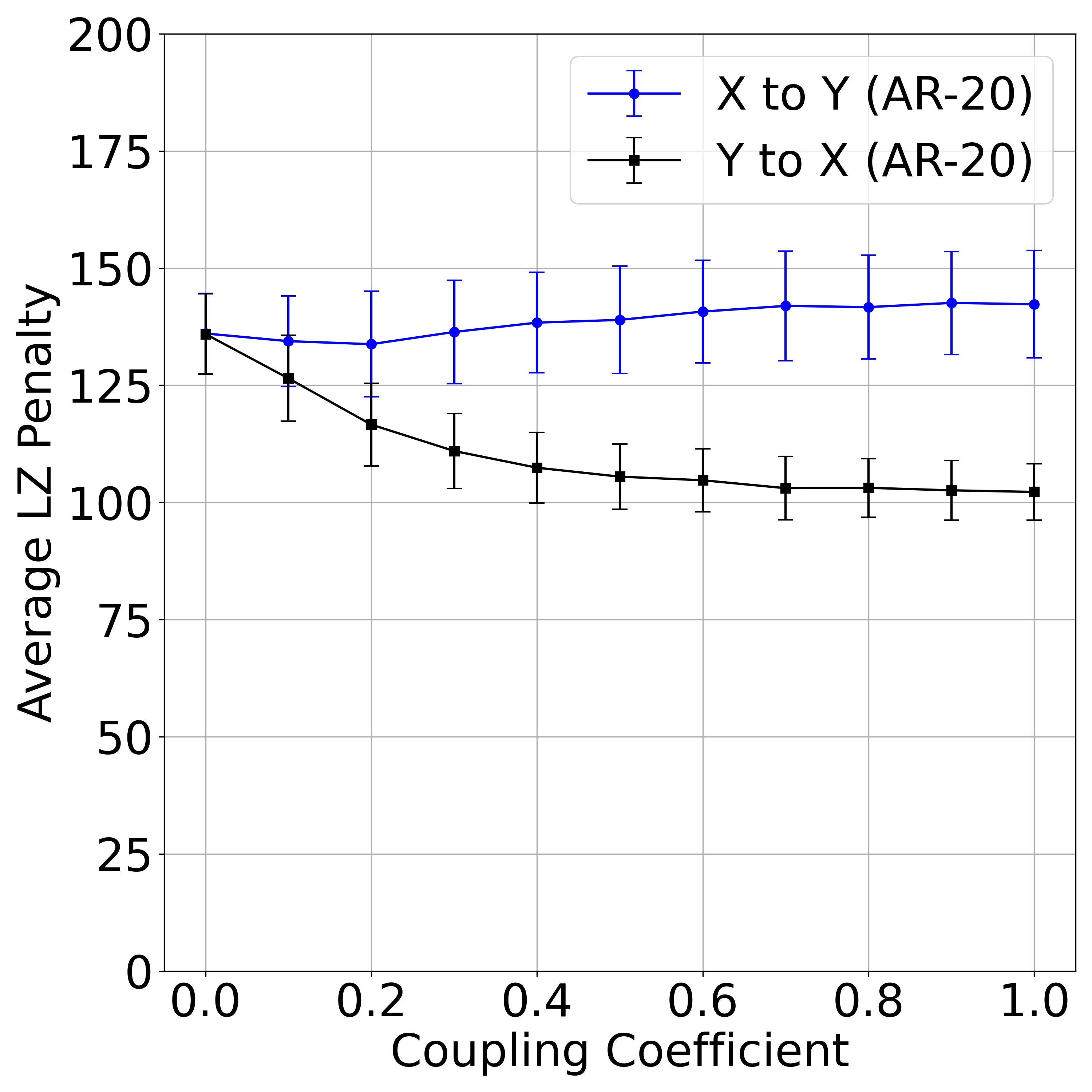}
        ~~(c)\label{fig_ar_20_var_LZ_penalty_coupling_coefficient}
    \end{subfigure}
    \hfill
    \begin{subfigure}[]{0.45\textwidth}
        \centering
        \includegraphics[width=\linewidth]{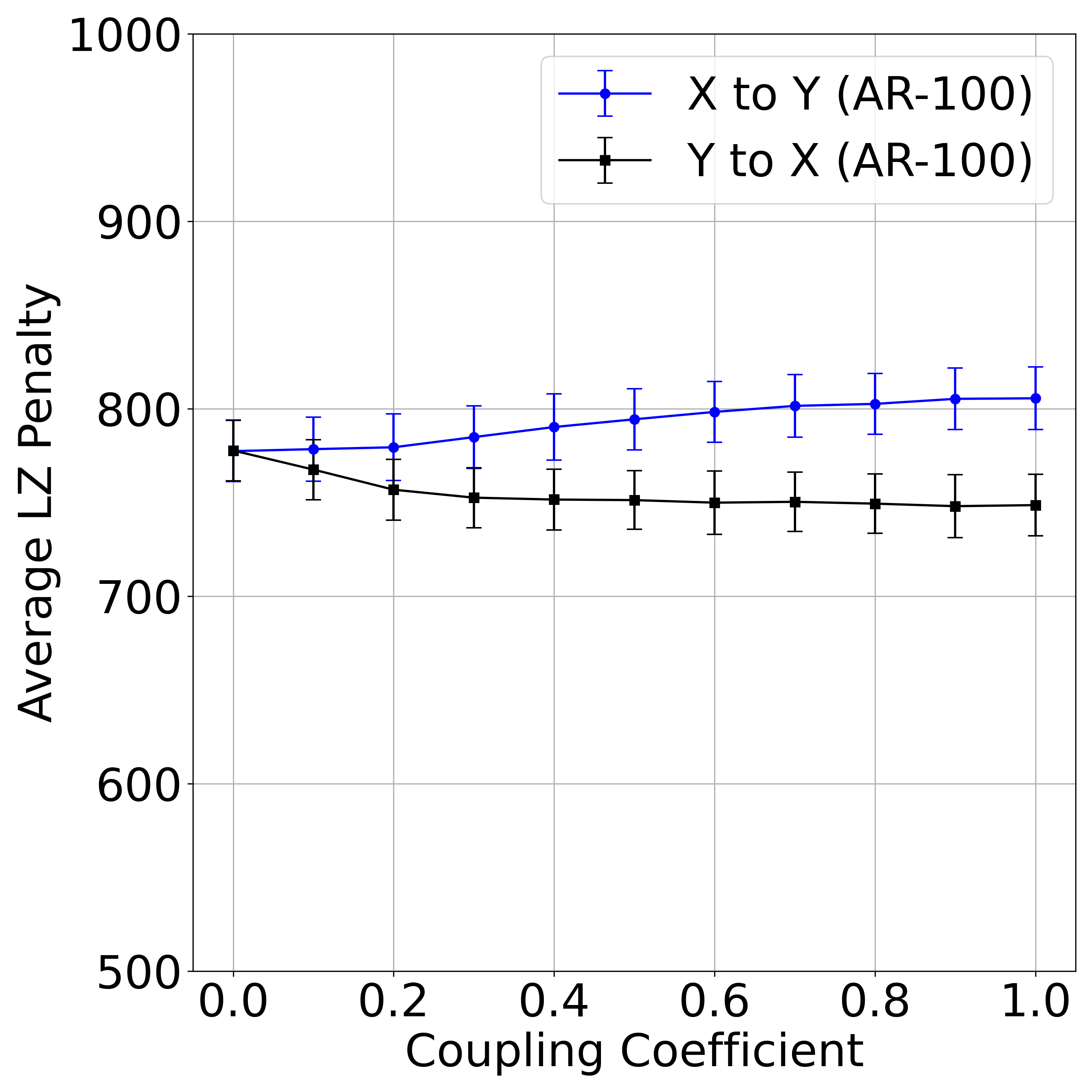}
        ~~(d)\label{fig_ar_100_var_LZ_penalty_coupling_coefficient}
    \end{subfigure}
    \caption{(a) Average LZ Penalty vs Coupling coefficients for AR-1 process. (b) Average LZ Penalty vs Coupling coefficients for AR-5 process. (c) Average LZ Penalty vs Coupling coefficients for AR-20 process. (d) Average LZ Penalty vs Coupling coefficients for AR-100 process. The coupling coefficients are varied from 0 to 1 with a step size of 0.1. For each coupling coefficient, LZ penalty was averaged across 1000 independent random trials.}  \label{fig_ar_1_5_20_100_var_LZ_penalty_coupling_coefficient}\end{figure}
\subsection{Coupled Logistic Map\label{sec:coupled_logistic_map}}
The 1D Logistic map is a chaotic map commonly used in population dynamics~\cite{may1976simple} study. The equations governing the dynamics of master-slave system of 1D Logistic map is as follows:
\begin{eqnarray}
    Y(t)=L_{1}(Y(t-1)),\\
    X(t) = (1-\eta)L_{2}(X(t-1)) + \eta Y(t-1).
\end{eqnarray}

The coupling coefficient $\eta$ is varied from $0$ to $0.9$ with a step size of $0.1$.  $L_{1}(t) = A_1 \cdot L_{1}(t-1)(1-L_{1}(t-1))$, and  $L_{2}(t) = A_2 \cdot L_{2}(t-1)(1-L_{2}(t-1))$, where $A_1=4$ and $A_2=3.82$.

For both systems, $1000$ data instances ($Y(t)$ cause , $X(t)$ (effect)) are generated. Each of the data instances are of length $2000$, after removing the initial $500$ samples (transients) from the time series. Figure~\ref{fig_logistic_map_LZ_penalty_coupling_coefficient} represents the LZ penalty from $X(t)$ to $Y(t)$ and $Y(t)$ to $X(t)$ averaged across 1000 independent random trials for each coupling coefficient. We observe that the proposed measure correctly identifies the causal direction upto $\eta =0.4$ as depicted in Figure~\ref{fig_logistic_map_LZ_penalty_coupling_coefficient}~(a). From $\eta = 0.4$ onwards the timeseries $X(t)$ and $Y(t)$ are synchronized yielding to similar $LZ$ penalty.

\begin{figure}[!ht] 
    \centering
    \begin{subfigure}[]{0.45\textwidth}
        \centering
         \includegraphics[width=\linewidth]{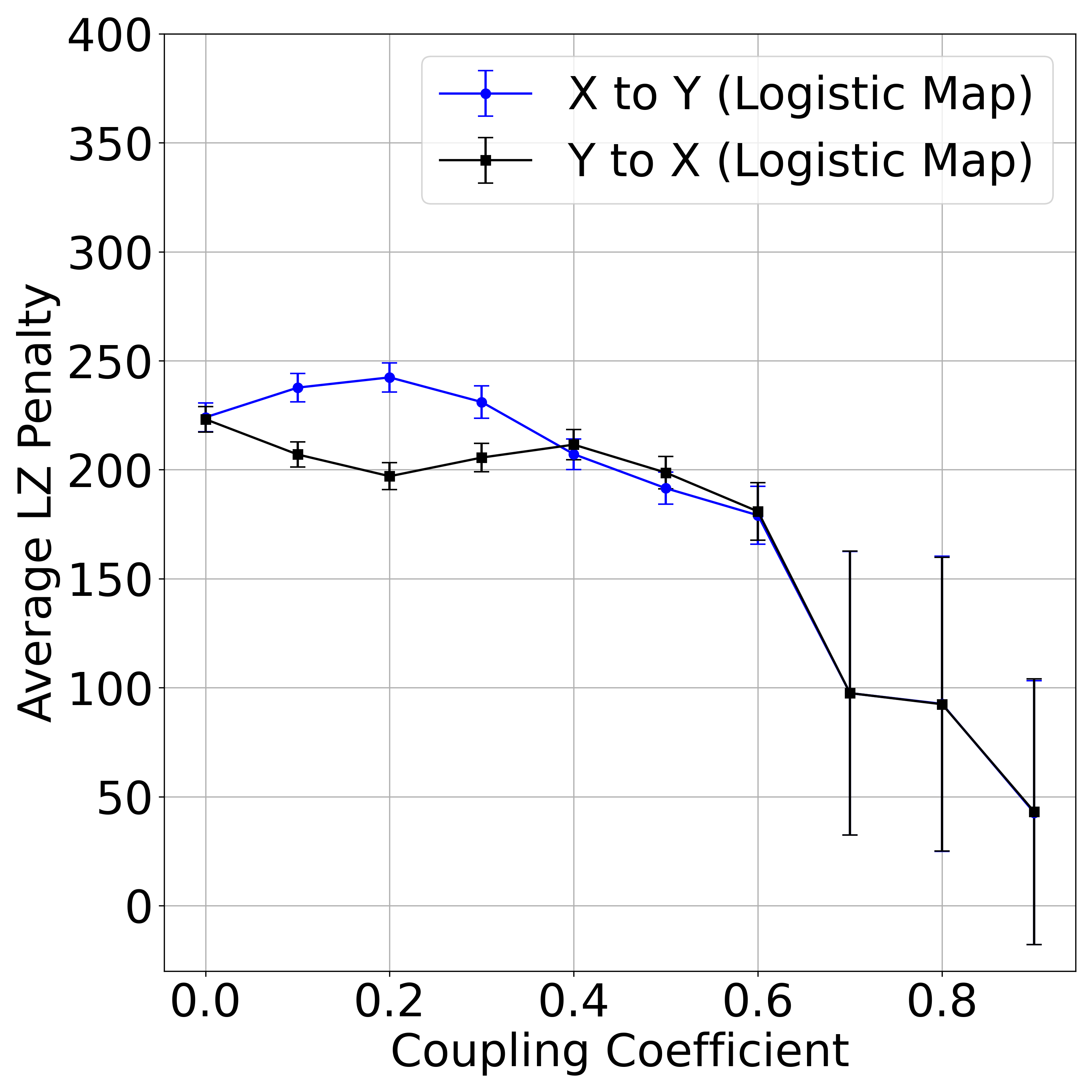}
         ~~(a)
     \end{subfigure}
    \hfill
    \begin{subfigure}[]{0.45\textwidth}
        \centering
        \includegraphics[width=\linewidth]{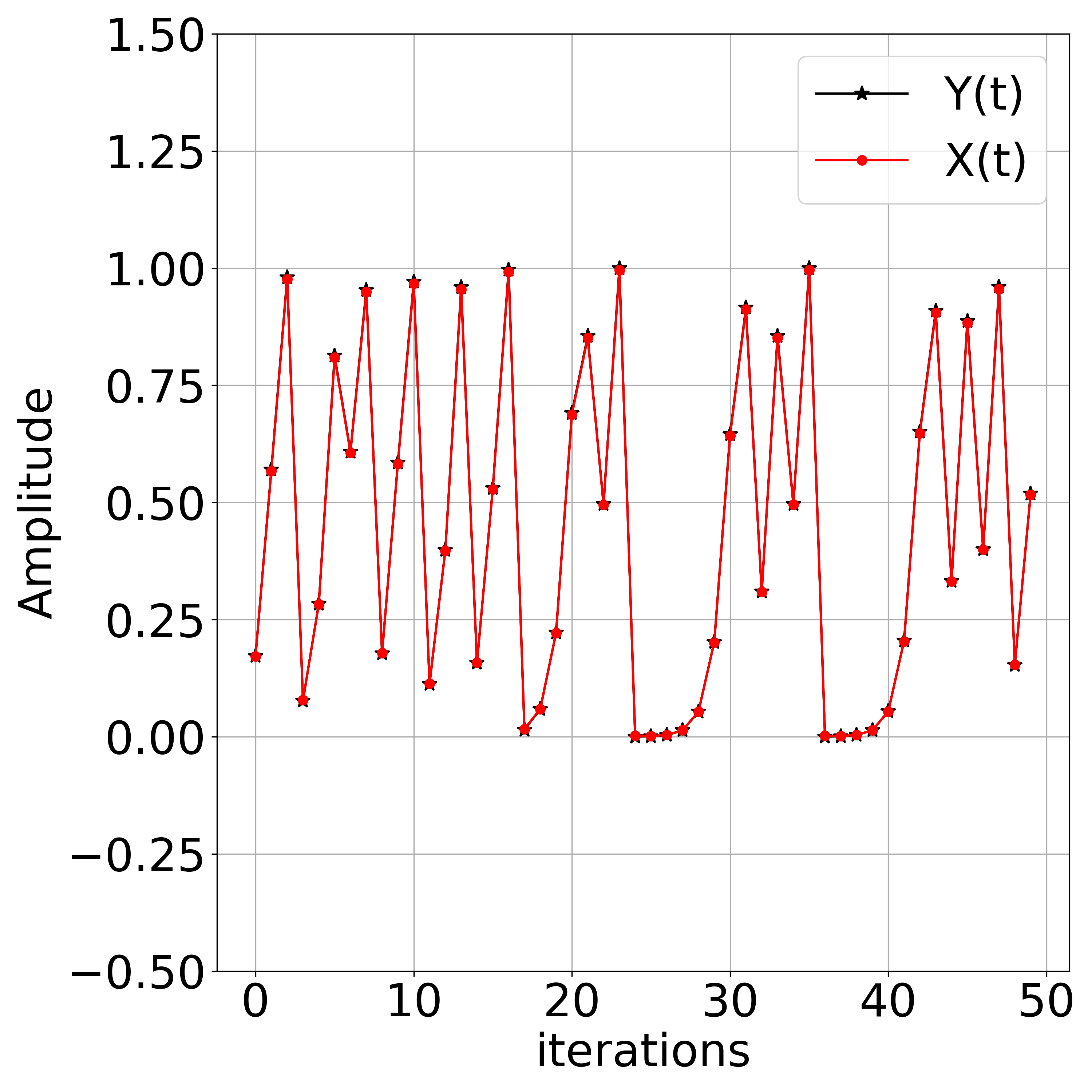}
        ~~(b)\label{fig_logistic_map_0.4_coupling_coefficient}
    \end{subfigure}
        \caption{(a) LZ Penalty vs Coupling coefficients for coupled chaotic logistic map averaged across 1000 independent trials for each coupling coefficient. The coupling coefficients are varied from 0 to 0.9 with a step size of 0.1. (b) An instance of the timeseries X(t) and Y(t) for coupling coefficient $\eta = 0.4$ for first 50 iterations, indicating synchronization between X(t) and Y(t).}
        \label{fig_logistic_map_LZ_penalty_coupling_coefficient}
    \end{figure}
 
\subsection{Tuebingen Cause Effect Pairs\label{sec:tubingen_cause-effect_pairs}}
Tuebingen data~\cite{mooij2016distinguishing},~\cite{UCI} consists of $108$ cause effect pairs. Out of which, we conducted our experiments on $104$ cause-effect pairs. For each cause effect pairs, the ground truth is provided. There are only two variables indicated as $X$ and $Y$. We use this data to determine whether $X\rightarrow Y$ or $Y\rightarrow X$ using the proposed $LZ$ penalty measure. We evaluate the efficacy of the problem in determining the causal direction by comparing with the given ground truth. We use macro F1-score and accuracy to evaluate the proposed measure's performance. In our experiments, we obtain a macro F1-score $= 0.44$, accuracy = $50.8\%$. The top $k$\% of pairs based on strength of causation is taken for $k = 1$ to $k = 100$, with a step size of $1$. Figure~\ref{fig_acc_vs_decision_rate_tubengen} provides the accuracy vs decison rate ($k$\%) of the proposed $LZ$ Penalty measure in comparison with $LZ-P$ measure defined in~\cite{pranay2021causal} (Note: the proposed formulation in this research is completely different from the $LZ-P$ formulation defined in~\cite{pranay2021causal}). The research in~\cite{budhathoki2018origo} have reported an accuracy of approximately $58\%$ for a selected $95$ datasets from Tuebingen Cause-Effect data pairs. 

\begin{figure}[!ht]\centering \includegraphics[width=0.5\textwidth]{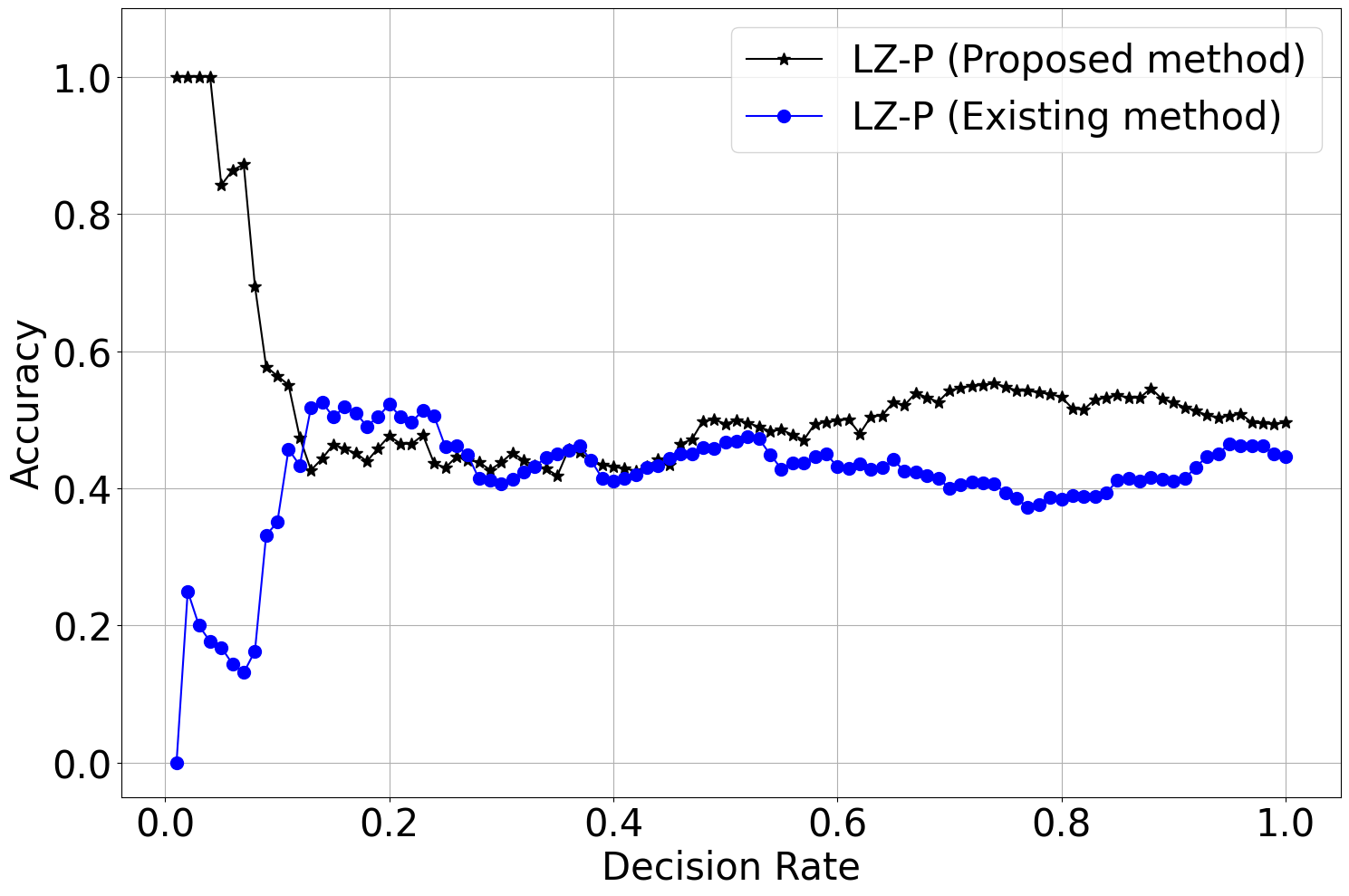} \caption{Accuracy vs. Decision rates for Tuebingen dataset using proposed LZ Penalty measure and LZ-P measure defined in~\cite{pranay2021causal}.} \label{fig_acc_vs_decision_rate_tubengen} \end{figure}
\newpage
\subsection{Incorporation of the proposed causal and distance based measure in decision tree}

In this section, we compare the performance evaluation of the proposed decision tree using $LZ$ based causality measure, $LZ$ based distance metric with Decision tree using gini impurity. The dataset used and their train-test distribution are provided in Table~\ref{table:test-train split}.
\begin{table}[!ht]
\centering
\caption{Train-test split information for each dataset. 80\%-20\% train-test ratio was utilized. Imbalance Ratio is defined as the ratio of number of training samples in the most commonly occurring class to the number of training samples in the least commonly occurring class. It is calculated for the training data.}
\resizebox{\textwidth}{!}{%
\begin{tabular}{|l|l|l|l|l|l|}
\hline
\textbf{Dataset} &
  \textbf{Classes} &
  \textbf{Features} &
  \textbf{\begin{tabular}[c]{@{}l@{}}Training \\ samples\\/class\end{tabular}} &
  \textbf{\begin{tabular}[c]{@{}l@{}}Testing \\ samples\\/class\end{tabular}} & \textbf{\begin{tabular}[c]{@{}l@{}}Imbalance \\ Ratio of \\ train data\end{tabular}}\\ \hline
\begin{tabular}[c]{@{}l@{}}Iris~\cite{scikit-learn}; \\ \cite{iris_original}  \end{tabular}                                                              & 3 & 4  & (39,37,44) & (11,13,6) & 1.189 \\ \hline

\begin{tabular}[c]{@{}l@{}}Breast Cancer Wisconsin\\~\cite{UCI} ; \\\cite{bcwd_original}\end{tabular}  & 2  & 30   & (286,169) & (71,43)    & 1.692         \\ \hline
\begin{tabular}[c]{@{}l@{}}Congressional\\ Voting Records\\~\cite{UCI, congressional_voting_records_105}\end{tabular} & 2 & 16  & (101, 84)   & (23,14)  & 1.202 \\ \hline
\begin{tabular}[c]{@{}l@{}}Car Evaluation\\~\cite{UCI};\\ ~\cite{ car_evaluation_19}    \end{tabular}                                                            & 4 & 6 & (307, 55, 968,  52) &  (77, 14, 242, 13) & 18.615 \\ \hline

\begin{tabular}[c]{@{}l@{}}KRKPA7 ~\cite{UCI};\\ \cite{chess_(king-rook_vs._king-pawn)_22}   \end{tabular}                                                  & 2 & 35 & (1227, 1329)   & (300, 340)  & 1.083   \\ \hline
\begin{tabular}[c]{@{}l@{}}Mushroom ~\cite{UCI};\\ \cite{ mushroom_73} \end{tabular}                                                    & 2 & 22 & (2783, 1732)   & (705, 424)  & 1.607   \\ \hline
\begin{tabular}[c]{@{}l@{}}Heart Disease\\~\cite{UCI}; \\ \cite{ heart_disease_45}   \end{tabular}                                                            & 5 & 13  & (124, 45, 30, 28, 10) & (36, 9, 5, 7, 3) & 12.4 \\ \hline
\begin{tabular}[c]{@{}l@{}}Thyroid~\cite{UCI};\\ \cite{thyroid_disease_102}    \end{tabular}                                                           & 3 & 21  & (93, 191, 3488) & (73,  177, 3178) & 37.505 \\ \hline
AR Dataset                                                     & 2 & 1  & (113, 127) & (25, 35) & 1.123 \\ \hline
\end{tabular}%
}
\label{table:test-train split}
\end{table}
\subsubsection{Results and Discussion}
This section describes the experimental results carried out to test the efficacy of $LZ$ distance metric based decision tree and $LZ$ causal metric based decision tree. We test the model's performance on the datasets described in Appendix section~\ref{sec:datasets}, along with a synthetic autoregressive causal dataset that we generate, termed as the AR dataset,
This dataset has been synthetically generated to provide data where there is an underlying causal structure between the various training examples. 450 training samples of an AR(1) process were generated according to the following equations: 
  \begin{eqnarray}
      X(t) = aX(t-1) + \eta Y(t-1) + \epsilon_{X,t} \ \\ Y(t) = bY(t-1) + \epsilon_{Y,t}
  \end{eqnarray}
 
Coupling coefficient $(\eta)$ was set to be 0.7, and parameters were fixed as $a$ = 0.8, $b$ = 0.8. Initial values of $X$ and $Y$ were set to 0 and the noise terms $ \epsilon_{X,t}$ and $\epsilon_{Y,t}$ were modeled as $\nu ~e_1$ and $\nu ~e_2$ where $e_1$ and $e_2$ were sampled from the standard normal distribution and the noise intensity $ \nu$=0.03. The timeseries $Y$ was taken to be the feature data. The timeseries $X$ was taken as the target to be classified after equi-width binning. The first 150 transients were removed, and the remaining samples were used for classification.\\
The train-test split for the datasets is done as described in Table~\ref{table:test-train split}. 
The hyperparameter tuning for $LZ$ distance metric based decision tree and $LZ$ causal metric based decision tree of the aforementioned datasets are mentioned in Appendix section~\ref{sec:appendix_hyperparemeter}. The summary of the results of the two proposed models are highlighted in Table~\ref{tab:performance} in Appendix section~\ref{sec:appendix:accuracies}, along with the performance of a decision tree that uses Gini impurity as the splitting criteria. The comparison between the three models is illustrated in Figure~\ref{fig:barplotf1}.

\begin{figure}[!ht] \centering \includegraphics[width=0.8\textwidth]{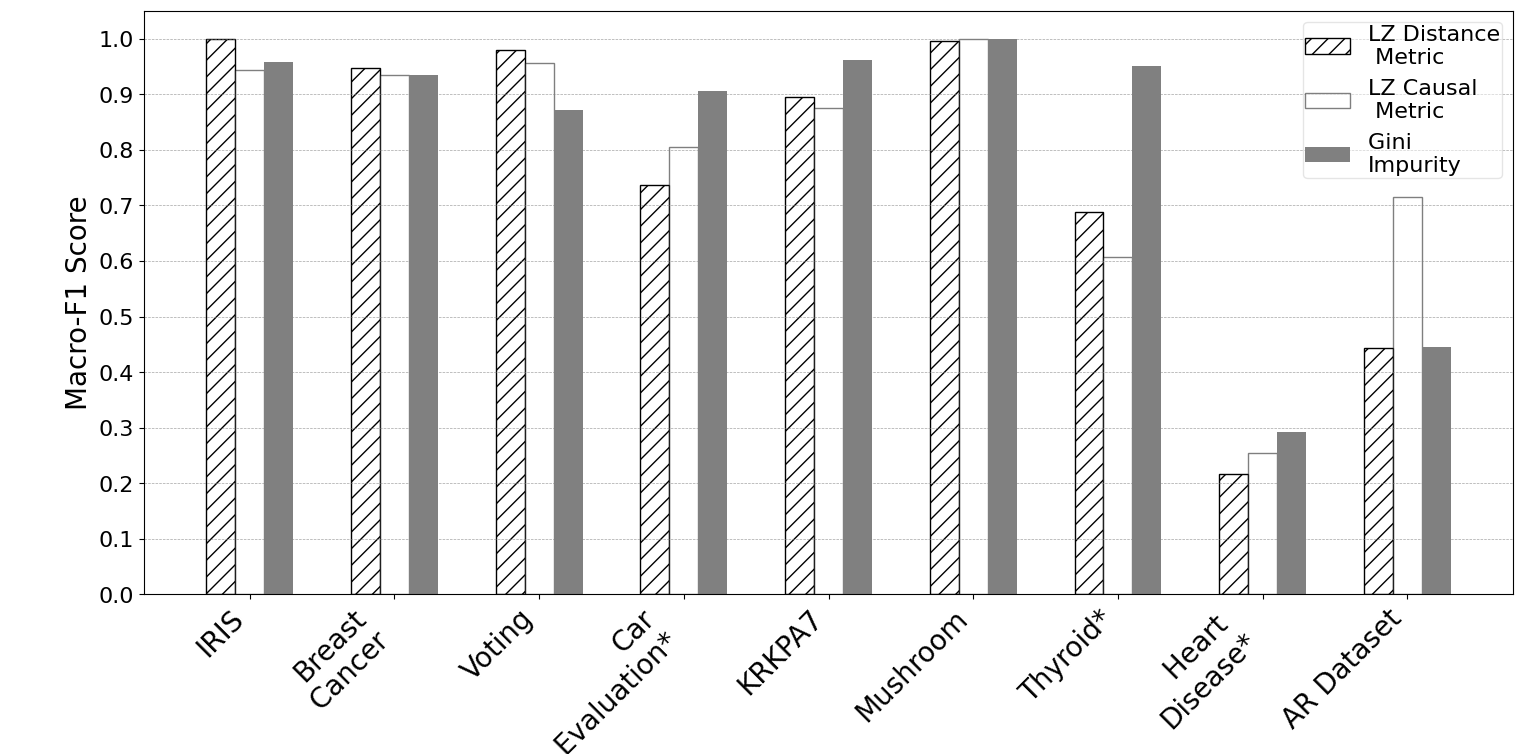} \caption{Bar Graph of macro F1 scores of predictions made by LZ distance metric based decision trees, LZ causal metric based decision trees and Gini impurity based decision trees, for various datasets. The datasets marked with a `*' are highly imbalanced. } \label{fig:barplotf1} \end{figure}
The macro F1 scores obtained by the three models are comparable across most balanced datasets. 
The macro F1 score of the $LZ$-causal metric based decision tree on the AR dataset is 60.5\% higher than both the macro F1 scores obtained by the other two algorithms on the same dataset. This result suggests that the causal decision tree effectively captures the intrinsic causal and temporal structures within the data, outperforming other models in discerning these underlying relationships. 
The Gini-entropy based decision tree significantly outperforms the Lempel Ziv complexity based models when the data is imbalanced, as seen in the Thyroid and Car evaluation datasets.

Table~\ref{table:causal-strength-ranking-heart} and Table~\ref{table:causal-strength-ranking-mushroom} (Appendix Section~\ref{sec:appendix:ranking_mushrom_dataset})  illustrate the ranking for features on the Heart Disease and Mushroom datasets respectively.

\begin{figure}[!ht] \centering \includegraphics[width=0.8\textwidth]{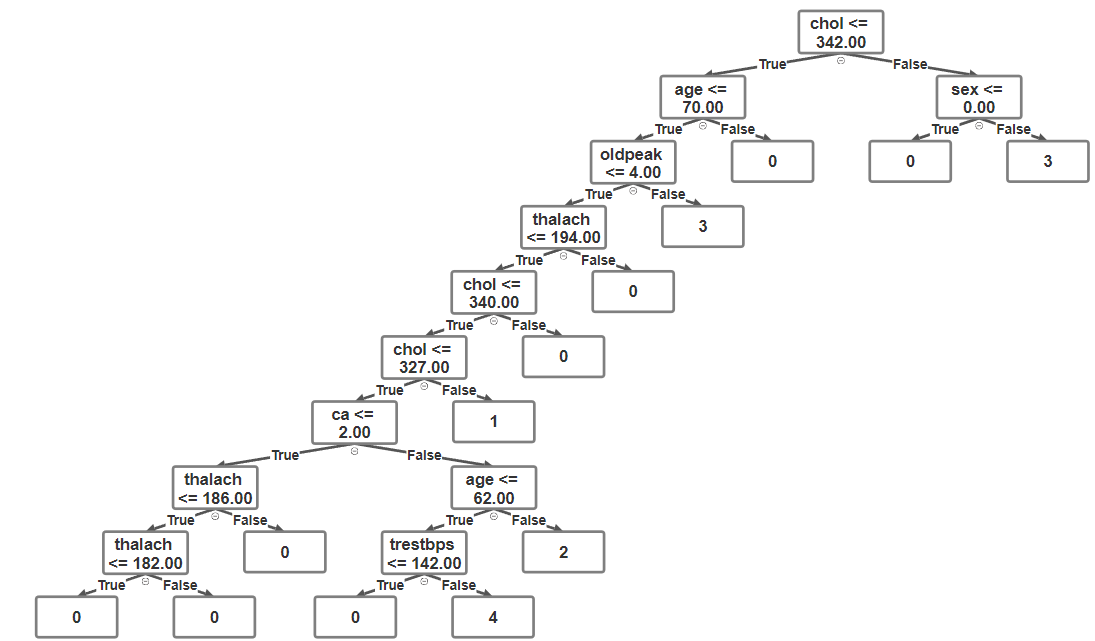}
\parbox{\linewidth}{\footnotesize\textit{}Abbreviations used: \textit{chol}-Serum Cholesterol in mg/dl, \textit{oldpeak}-ST depression induced by exercise relative to rest, \textit{thalach}-Maximum heart rate achieved,\textit{ca}-Number of major vessels (0-3) colored by fluoroscopy, \textit{trestbps}-Resting blood pressure. }
\caption{Causal Decision Tree for \textit{Heart Disease} dataset.} \label{fig:causal_dt_heart_dat} 

\end{figure}

The limitations of this research are highlighted in Appendix Section~\ref{sec:appendix:limitations}. We consider this as an opportunity to further develop the proposed method.

\begin{table}[!ht]
    \centering
    \caption{Ranking of Features of the Heart Disease dataset based on Causal Strength to the presence of heart disease.}
    \begin{tabular}{|l|c|}
        \hline
        \textbf{Feature} & \textbf{Causal Strength}  \\ \hline
        Serum Cholesterol in mg/dl & 0.4361 \\ \hline
        Age & 0.2025  \\ \hline
        Sex & 0.1994 \\ \hline
        ST depression induced by exercise relative to rest & 0.0997  \\ \hline
        Maximum heart rate achieved
 & 0.0545  \\ \hline
        Number of major vessels (0-3) colored by fluoroscopy
 & 0.0062  \\ \hline
          Resting blood pressure
 & 0.0016  \\ \hline
  
    \end{tabular}
    \label{table:causal-strength-ranking-heart}
\end{table}

\newpage
\section{Conclusion\label{sec:conclusion}}
In this research, we propose a novel causality testing measure and distance metric based Lempel-Ziv complexity. Using the causality based measure, we were able to successfully capture the direction of causation of coupled AR processes and coupled logistic maps. We then incorporated the $LZ$ based causality measure in decision trees to build trees that makes use of a causally informed splitting criteria. The efficacy of $LZ$ distance metric based decision tree was also evaluated in the research experiments. We further provide a metric to analyze the feature importance based on causal strength derived from the $LZ$ based causal decision tree. In future work, we will test the efficacy of the model on temporal causal data. We also plan to explore the applicability of the proposed distance metric based on Lempel-Ziv complexity in the bio-informatics domain, where order of nucleotide sequence plays a role. 



\section{Appendix\label{sec:appendix}}
\subsection{Appendix A: Proof of $d_{LZ}(G_x,G_y)$ being a distance metric}\label{sec:appendix_proof}
The proof of $d_{LZ}(G_x,G_y)$ being a distance metric is as follows.\\
Consider two strings $x$ and $y$. Let $G_x$ and $G_y$ be the grammar encoded by the $LZ$ complexity algorithm for $x$ and $y$ respectively.
\subsubsection*{Definition of the Distance Function}
Let $G_x$ and $G_y$ be two sets. Define the distance between them as:
\[
d_{LZ}(X,Y) = |G_x / G_y| + |G_y / G_x|
\]
where $|A|$ denotes the cardinality (size) of a set $A$. We will show that $d_{LZ}(G_x,G_y)$ satisfies the four properties of a metric.

\subsubsection*{1. Non-Negativity}
We need to show that $d_{LZ}(G_x,G_y) \geq 0$ for all sets $G_x$ and $G_y$. 

By definition, $G_x / G_y$ and $G_y / G_x$ are both sets, and the cardinality of any set is non-negative. Hence:
\[
|G_x / G_y| \geq 0 \quad \text{and} \quad |G_y / G_x| \geq 0
\]
Thus,
\[
d_{LZ}(G_x, G_y) = |G_x / G_y| + |G_y / G_x| \geq 0
\]
Non-negativity is satisfied.

\subsubsection*{2. Identity of Indiscernibles}
We must show that $d_{LZ}(G_x,G_y) = 0$ if and only if $G_x = G_y$.

\textbf{If $G_x = G_y$:}
If $G_x = G_y$, then:
\[
G_x - G_y = \varnothing \quad \text{and} \quad G_y - G_x = \varnothing
\]
Thus:
\[
|G_x / G_y| = 0 \quad \text{and} \quad |G_y / G_x| = 0
\]
Therefore, $d_{LZ}(G_x, G_y) = 0$.

\textbf{If $d_{LZ}(G_x, G_y) = 0$:}
If $d_{LZ}(G_x, G_y) = 0$, then, as both are individually greater than or equal to zero:
\[
|G_x / G_y| = 0 \quad \text{and} \quad |G_y / G_x| = 0
\]
This implies that:
\[
G_x - G_y = \varnothing \quad \text{and} \quad G_y - G_x = \varnothing
\]
which means $G_x = G_y$.

Thus, identity of indiscernibles is satisfied.

\subsubsection*{3. Symmetry}
We need to show that $d_{LZ}(G_x, G_y) = d_{LZ}(G_y, G_x)$.

By the definition of set difference:
\[
G_x / G_y = \{ x \in G_x \mid x \notin G_y \}
\]
and
\[
G_y / G_x = \{ y \in G_y \mid y \notin G_x \}
\]
Therefore:
\[
d_{LZ}(G_x, G_y) = |G_x / G_y| + |G_y / G_x|
\]
and
\[
d_{LZ}(G_y, G_x) = |G_y / G_x| + |G_x / G_y|
\]
Since addition is commutative, we have:
\[
d_{LZ}(G_x, G_y) = d_{LZ}(G_y, G_x)
\]
Thus, symmetry is satisfied.

\subsubsection*{4. Triangle Inequality}
We need to show that for any sets $G_x$, $G_y$, and $G_z$:
\[
d_{LZ}(G_x, G_z) \leq d_{LZ}(G_x, G_y) + d_{LZ}(G_y, G_z)
\]
This is equivalent to showing:
\[
|G_x / G_z| + |G_z / G_x| \leq \left( |G_x / G_y| + |G_y / G_x| \right) + \left(|G_y / G_z| + |G_z / G_y| \right)
\]

Consider some element in the set $G_x / G_z$. It either belongs to $G_x / G_y$, or $G_y / G_z$. This is true for every element in the set. Therefore, we have:
\[
|G_x / G_z| \leq |G_x / G_y| + |G_y / G_z|
\]
and similarly, for $G_z / G_x$,
\[
|G_z / G_x| \leq |G_z / G_y| + |G_y / G_x|
\]
Adding these inequalities gives:
\[
|G_x / G_z| + |G_z / G_x| \leq \left( |G_x / G_y| + |G_y / G_x| \right) + \left(|G_y / G_z| + |G_z / G_y| \right)
\]
Thus, the triangle inequality is satisfied.

\subsubsection*{Conclusion}
Since $d_{LZ}(G_x, G_y)$ satisfies non-negativity, identity of indiscernibles, symmetry, and the triangle inequality, it is a valid distance metric on the space of set of all sets.
\subsection{Appendix B: Datasets} \label{sec:datasets}
\begin{enumerate}
    \item IRIS: \\
    This dataset consists of 150 instances. There are three labels corresponding to three different types of irises', namely 'setosa', 'versicolour' and 'virginica'. The features describe the sepal length, sepal width, petal length and petal width measured in centimeters.
 \item Breast Cancer Wisconsin \\
 This dataset contains features derived from cell images to classify breast cancer as either benign or malignant. There are 569 samples with 30 numerical features describing characteristics of cell nuclei.
 \item Congressional Voting Dataset \\
 This dataset contains the information of the votes of each U.S House of Representatives Congressman on 16 key issues. Each of the 435 training samples is labeled by party affiliation, 'republican' or 'democrat'.
 \item Car Evaluation \\
A classification dataset with 1,728 instances that classify cars as unacceptable, acceptable, good, or very good based on six categorical attributes: buying price, maintenance cost, number of doors, capacity, size of luggage boot, and safety.
 \item KRKPA7 \\
  This dataset, also known as ``King-Rook-King-Pawn-Against-Seven", is a chess dataset with 3196 samples. The 35 attributes describe the chess board in an endgame, and the training samples are classified based on whether White can win.
 \item Mushroom \\
 he Mushroom dataset contains records for 8,124 mushrooms, each classified as edible or poisonous. The dataset has 22 categorical attributes describing physical characteristics like cap shape, color, and odor, across 23 different species.
 \item Heart Disease  \\
 This dataset is used for predicting heart disease presence, with 303 samples and 13 features such as age, sex, cholesterol levels, and blood pressure. The label indicates disease presence (1,2,3,4) or absence (0).
 \item Thyroid \\
 The Thyroid dataset focuses on thyroid disease detection, with instances labeled for three classes: normal, hyperthyroid, and hypothyroid. There are 3772 training samples and 3428 testing samples, over 21 features.

\end{enumerate}
\subsection{Appendix C: Hyperparameter tuning\label{sec:appendix_hyperparemeter}}
In this section, the hyperparameter tuning for $LZ$ distance metric based decision tree and $LZ$ causal metric based decision tree are elaborated. For $LZ$ distance metric based decision tree, we used stratified five fold crossvalidation using random split. In the case of $LZ$ causal metric based decision tree, as well as the Gini entropy based Decsion tree, we used five fold crossvalidation using timeseries split. Table~\ref{tab_hyperparameter_tuning} provides the best hyperparameters obtained after crossvalidation experiments for $LZ$ distance metric based decision tree and $LZ$ causal metric based decision tree corresponding to the datasets considered in this research.
Hyperparameter tuning was done taking a range of 1 to 10 for the minimum number of samples per node, with a step size of 1. Maximum depth of tree was taken from 1 to 20 with the step size of 1. 

\begin{table}[!h]
    \centering
        \caption{Tuned Hyperparameters and Training Performance Metrics for LZ-Distance-Metric based Tree, LZ-Causal-Metric Based Tree and Gini Impurity Based Tree across datasets}
    \begin{adjustbox}{max width=\textwidth}
    \begin{tabular}{|c|c|c|c|c|c|}
        \hline
        \textbf{Datasets} & {}&\textbf{Min Samples} & \textbf{Max Depth} & \textbf{Average F1 score on train data} & \textbf{Variance of F1 Score} \\ \hline
        
        \multirow{3}{*}{Iris} & LZ Distance Metric & 2 & 11 & 0.94 & 0.004\\ \cline{2-6}
                                   & LZ Causal Metric   & 9 & 6 & 0.89 & 0.016 \\ \cline{2-6}
                                    & Gini Impurity & 9 & 4 & 0.953 & 0.002 \\ \hline
                                   
        \multirow{2}{*}{Breast Cancer Wisconsin} &  LZ Distance Metric & 9 & 4 & 0.89 & 0.0006 \\ \cline{2-6}
                                   &  LZ Causal Metric  & 2 & 6 & 0.91 & 0.0007 \\ \cline{2-6}
                                      & Gini Impurity & 4 & 8 & 0.931 & 0.0008 \\ \hline
                                   
        \multirow{2}{*}{Congressional Voting Dataset} & LZ Distance Metric & 7 & 4 & 0.930 & 0.002 \\ \cline{2-6}
                                   & LZ Causal Metric    & 5 & 9 & 0.965 & 0.0001 \\ \cline{2-6}
                                      & Gini Impurity & 9 & 9 & 0.979 & 0.0002 \\ \hline

        \multirow{2}{*}{Car Evaluation} & LZ Distance Metric& 2 & 14 & 0.66 & 0.005\\ \cline{2-6}
                                   & LZ Causal Metric & 2 & 9 & 0.514 & 0.0004\\ \cline{2-6}
                                      & Gini Impurity & 2 & 9 & 0.832 & 0.002 \\ \hline
                                   
         \multirow{2}{*}{KRKPA7} & LZ Distance Metric & 2 & 18 & 0.90 & 0.0002 \\ \cline{2-6}
                                   & LZ Causal Metric    & 4 & 9 & 0.862 & 0.002\\ \cline{2-6}
                                      & Gini Impurity & 3 & 9 & 0.968 & 0.0* \\ \hline
                                   
         \multirow{2}{*}{Mushroom} &LZ Distance Metric & 9 & 9 & 1.0 & 0.0* \\ \cline{2-6}
                                   & LZ Causal Metric    & 9 & 9 & 0.992 & 0.0*\\ \cline{2-6}
                                      & Gini Impurity & 9 & 7 & 1.0 & 0.0* \\ \hline
        
        \multirow{2}{*}{Heart Disease} & LZ Distance Metric & 6 & 19 & 0.226 & 0.0002 \\ \cline{2-6}
                                   &LZ Causal Metric   & 6 & 9 & 0.26 & 0.009 \\ \cline{2-6}
                                      & Gini Impurity & 3 & 9 & 0.301 & 0.002 \\ \hline
            \multirow{2}{*}{Thyroid } & LZ Distance Metric & 9 & 20 & 0.664 & 0.002 \\ \cline{2-6}
                                   &LZ Causal Metric   & 5 & 9 & 0.642 & 0.0005 \\ \cline{2-6}
                                      & Gini Impurity & 2 & 5 & 0.966 & 0.001 \\ \hline
            \multirow{2}{*}{AR Dataset } & LZ Distance Metric & 2 & 9 & 0.503 & 0.004 \\ \cline{2-6}
                                   &LZ Causal Metric   & 2 & 6 & 0.70 & 0.008 \\ \cline{2-6}
                                      & Gini Impurity & 2 & 6 & 0.55 & 0.004 \\ \hline
    \end{tabular}
    
\end{adjustbox}
\parbox{\linewidth}{\footnotesize\textit{} *Variance of the order of $10^{-5}$ or less which has been rounded off to 0. }
    \label{tab_hyperparameter_tuning}
\end{table}
\newpage
\subsection{Appendix D: Performance Metrics} \label{sec:appendix:accuracies}
This section contains the accuracies, F1 scores, precision score and recall score for $LZ$-Distance-Metric based Tree, $LZ$-Causal-Metric Based Tree and Gini-Entropy Based Tree acroos various datasets.
\begin{table}[!h]
    \centering
        \caption{Performance metrics for LZ-Distance-Metric based Tree, LZ-Causal-Metric Based Tree and Gini-Entropy Based Tree across datasets}
        \resizebox{\textwidth}{!}{%
    \begin{tabular}{|c|c|c|c|c|c|}
        \hline
        \textbf{Datasets} & \textbf{Algorithm}&\textbf{Accuracy} & \textbf{F1 Score} & \textbf{Precision} & \textbf{Recall} \\ \hline
        
        \multirow{3}{*}{Iris} & LZ Distance Metric based DT& 1.0 & 1.0 & 1.0 & 1.0 \\ \cline{2-6}
                                   & LZ Causal Metric based DT  & 0.933 & 0.943 & 0.956 & 0.939 \\ \cline{2-6} 
                                    & Gini Impurity based DT & 0.967 & 0.958 & 0.978 & 0.944 \\ 
                                   \hline
        \multirow{2}{*}{Breast Cancer Wisconsin} & LZ Distance Metric based DT & 0.947 & 0.948 & 0.939 & 0.943 \\ \cline{2-6}
                                   & LZ Causal Metric based DT   & 0.939 & 0.934 & 0.937 & 0.932 \\ \cline{2-6} 
                                    & Gini Impurity based DT & 0.938 & 0.934 & 0.936 & 0.932 \\ 
                                   \hline
                                   
        \multirow{2}{*}{Congressional Voting Records} & LZ Distance Metric based DT & 0.979 & 0.979 & 0.979 & 0.979 \\ \cline{2-6}
                                   & LZ Causal Metric based DT  & 0.957 & 0.957 & 0.96 & 0.958 \\ \cline{2-6} 
                                    & Gini Impurity based DT & 0.872 & 0.871 & 0.896 & 0.875 \\ 
                                   \hline
        \multirow{2}{*}{Car Evaluation} & LZ Distance Metric based DT& 0.829 & 0.682 & 0.734 & 0.664 \\ \cline{2-6}
                                   & LZ Causal Metric based DT  & 0.76 & 0.510 & 0.570 & 0.496 \\ \cline{2-6} 
                                    & Gini Impurity based DT & 0.945 & 0.883 & 0.914 & 0.865 \\ 
                                   \hline
         \multirow{2}{*}{KRKPA7 (Chess)} & LZ Distance Metric based DT & 0.895 & 0.895 & 0.895 & 0.896 \\ \cline{2-6}
                                   & LZ Causal Metric based DT  & 0.876 & 0.876 & 0.877 & 0.875 \\ \cline{2-6} 
                                    & Gini Impurity based DT & 0.961 & 0.961 & 0.960 & 0.962 \\ 
                                   \hline
        \multirow{2}{*}{Mushroom} & LZ Distance Metric based DT & 0.996 & 0.996 & 0.995 & 0.997 \\ \cline{2-6}
                                   & LZ Causal Metric based DT  & 1.0 & 1.0 & 1.0 & 1.0 \\  \cline{2-6} 
                                    & Gini Impurity based DT & 1.0 & 1.0 & 1.0 & 1.0 \\ 
                                   \hline
        \multirow{2}{*}{Heart Disease} & LZ Distance Metric based DT & 0.567& 0.217 & 0.210 & 0.235 \\ \cline{2-6}
                                   & LZ Causal Metric based DT  & 0.617 & 0.254 & 0.325 & 0.263 \\ \cline{2-6} 
                                    & Gini Impurity based DT & 0.55 & 0.292 & 0.293 & 0.308 \\ 
                                   \hline
           \multirow{2}{*}{Thyroid} & LZ Distance Metric based DT & 0.946& 0.688 & 0.827 & 0.640 \\ \cline{2-6}
                                   & LZ Causal Metric based DT  & 0.940 & 0.607 & 0.818 & 0.556 \\ \cline{2-6} 
                                    & Gini Impurity based DT & 0.992 & 0.951 & 0.946 & 0.957 \\ 
                                   \hline
         \multirow{2}{*}{AR Dataset} & LZ Distance Metric based DT & 0.467 & 0.444 & 0.444 & 0.445 \\ \cline{2-6}
                                   & LZ Causal Metric based DT  & 0.717 & 0.716 & 0.719 & 0.718 \\ \cline{2-6} 
                                    & Gini Impurity based DT & 0.45 & 0.446 & 0.45 & 0.448 \\ 
                                   \hline
         \end{tabular}%
        }

    \label{tab:performance}
\end{table}
\newpage
\subsection{Appendix E: LZ Distance Metric based decision tree and Gini entropy based decision tree}
The following section contains the LZ distance metric based decision tree and the Gini entropy based decision tree for the heart disease dataset, seen in Figure~\ref{fig:distance:heart} and Figure~\ref{fig:scikit:heart} respectively.
\begin{figure}[!ht]\centering \includegraphics[width=\textwidth]{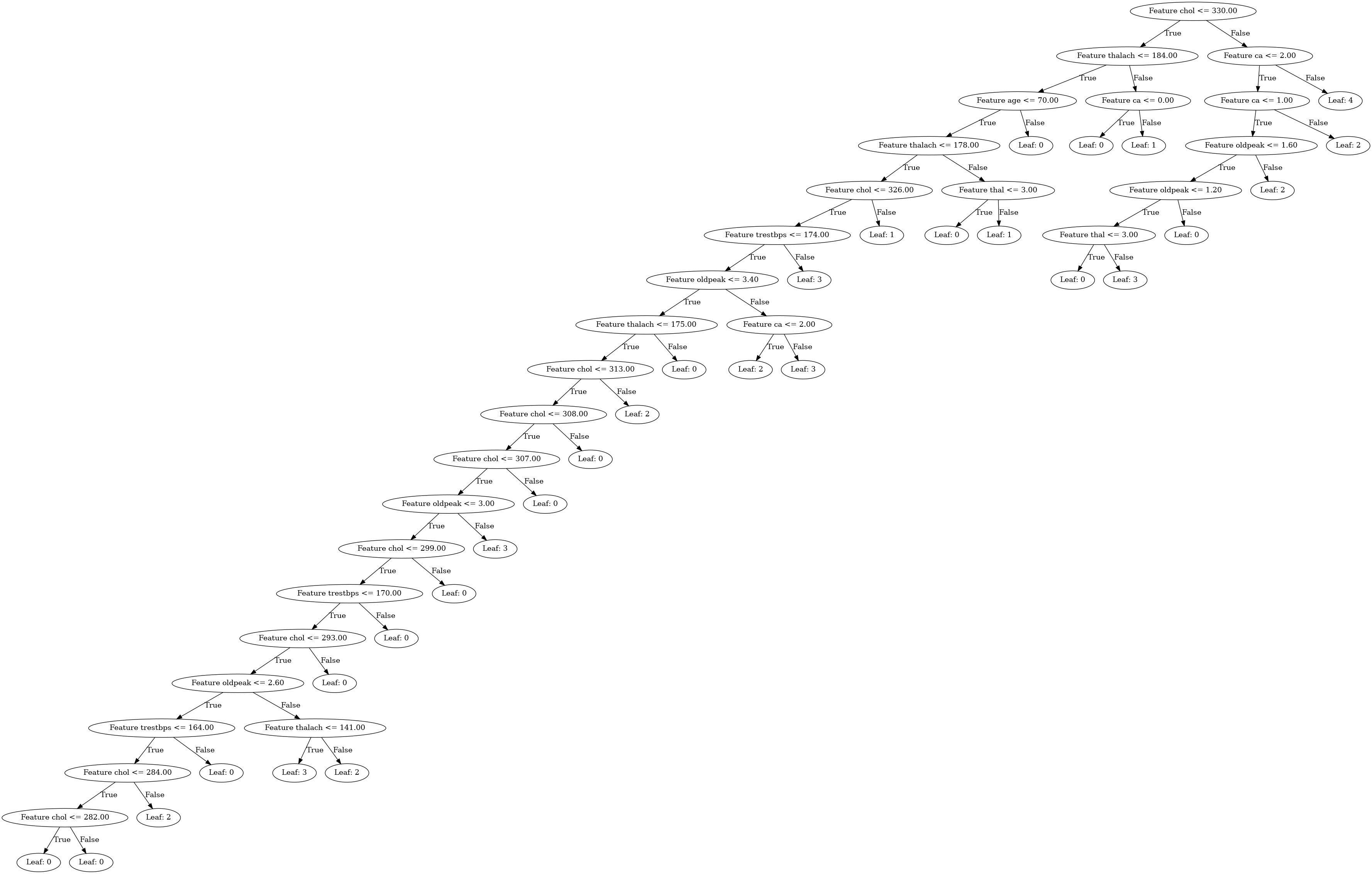} \caption{LZ Distance metric based decision tree generated for the heart disease dataset.} \label{fig:distance:heart} \end{figure}

\begin{figure}[!ht]\centering \includegraphics[width=\textwidth]{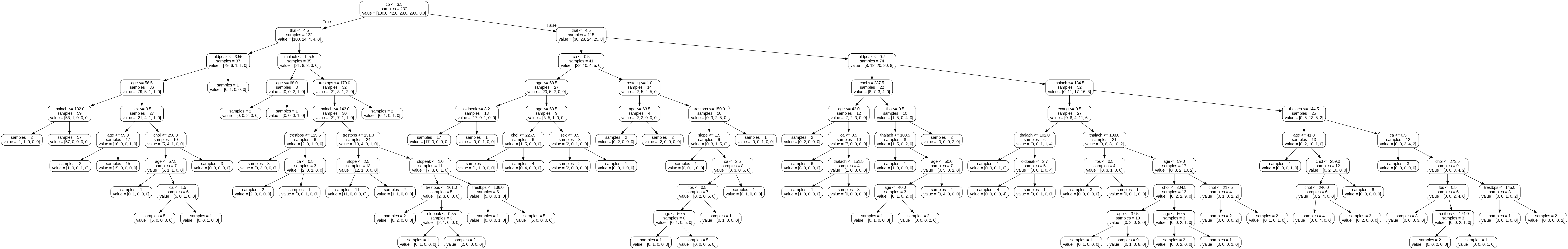} \caption{Gini Entropy based decision tree generated for the heart disease dataset.} \label{fig:scikit:heart} \end{figure}
\newpage
\subsection{Appendix F: Feature Ranking for Mushroom Dataset} 
\label{sec:appendix:ranking_mushrom_dataset}
This section contains the causal strength ranking for the features of the Mushroom dataset, seen in Table~\ref{table:mushroom}.
\begin{table}[h!]

    \centering
    \caption{Ranking of Features of the Mushroom dataset based on Causal Strength to whether the mushroom is poisonous.}\label{table:mushroom}
    \begin{tabular}{|l|c|}
        \hline
        \textbf{Feature} & \textbf{Causal Strength}  \\ \hline
        Odor & 0.3181   \\ \hline
        Cap Surface & 0.3041 \\ \hline
        Spore print color & 0.2444  \\ \hline
        Ring Number & 0.0374  \\ \hline
        Gill color
 & 0.0374  \\ \hline
        Habitat
 & 0.0187  \\ \hline
          Stalk color above ring
 & 0.0187  \\ \hline
   Stalk surface above ring
 & 0.0094  \\ \hline
  Stalk shape
 & 0.0094  \\ \hline
Cap colour
 & 0.0023  \\ \hline
    \end{tabular}
    \label{table:causal-strength-ranking-mushroom}
\end{table}

\newpage
\subsection{Appendix G: Limitations} \label{sec:appendix:limitations}
\begin{enumerate}
\item We consider an AR(1) process generated using the following equations: \begin{eqnarray} X(t) = aX(t-1) + \eta Y(t-1) + \epsilon_{X,t} \ \\Y(t) = bY(t-1) + \epsilon_{Y,t} \end{eqnarray}

For a coupling coefficient $\eta = 0$, we found that the measure is sensitive to the parameter $a$ (the coefficient of $X(t-1)$). We fixed the parameter $b = 0.6$ (the coefficient of $Y(t-1)$) and varied $a$ from 0.1 to 0.9, with a step size of 0.1. When using 2 bins to generate the symbolic sequence, the measure showed a spurious causal relationship. We hypothesize that this may be due to the binned sequence failing to capture the underlying dynamics of the real-valued data, leading to arbitrary causal inferences. Figure~\ref{fig_limitation_sensitivity_expts} illustrates the sensitivity observed using the proposed measure.

\begin{figure}[!ht] \centering \includegraphics[width=0.35\textwidth]{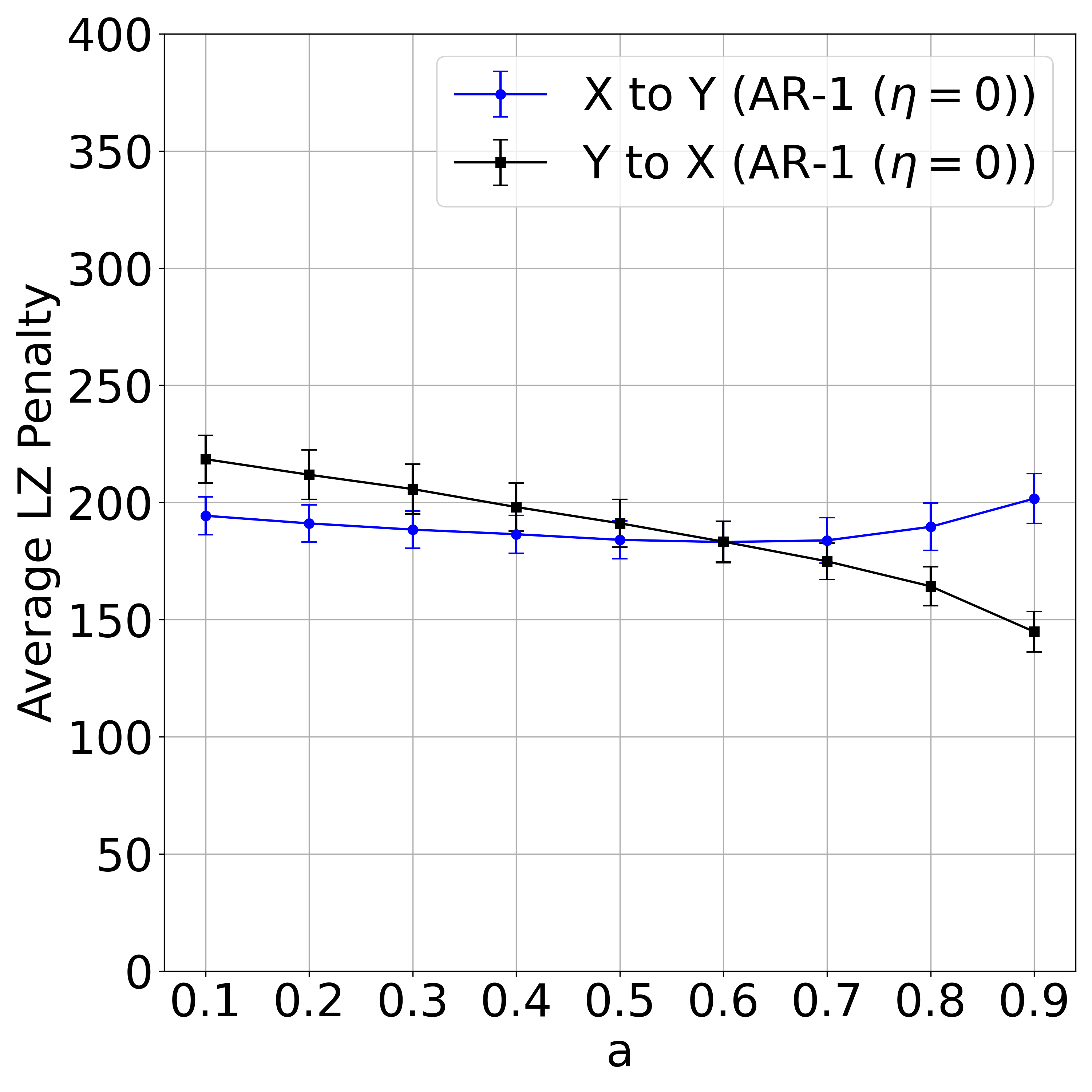} \caption{LZ Penalty vs. Coupling coefficient for the AR(1) process, averaged across 1000 independent trials with $\eta = 0$. The coefficient $a$ (of $X(t-1)$) is varied from 0.1 to 0.9 in increments of 0.1.} \label{fig_limitation_sensitivity_expts} \end{figure}

\item The decision trees that utilises LZ based Causal Measure as well as the LZ-based distance metrics are unable to capture the dynamics of imbalanced data. This may occur as both algorithms minimize either the Lempel Ziv Penalty between the symbolic sequence representation of the feature and the symbolic sequence representation of the target or the edit distance between the grammar sets of the same. In case of data imbalance, it may achieve this minimization by collapsing the feature representation into a homogeneous sequence of zeros, or otherwise fail to capture the appropriate thresholding.
   \item Validity of the proposed causal decision tree model: At present the validity of the proposed causal decision tree model relies on the interpretation of the domain expert. In the future work, we plan to investigate the notion of causal interpretability of the proposed decision tree model from an algorithmic point of view.

\end{enumerate}
\subsection{Appendix H: Code Availability}
The python code used in this research are available in the following GitHub repository: \url{https://github.com/i-to-the-power-i/causal-lz-p-decision-tree}.

\end{document}